\documentclass[letterpaper, 10 pt, conference]{ieeeconf}  

\IEEEoverridecommandlockouts                              

\usepackage{url}
\usepackage{graphics} 
\usepackage{graphicx} 
\usepackage{mathptmx} 
\usepackage{times} 
\usepackage{amsmath} 
\usepackage{amssymb}  
\usepackage{color}
\usepackage{subfig}
\usepackage{epstopdf}
\usepackage{balance}
\usepackage{wrapfig}
\usepackage{adjustbox}
\usepackage{booktabs, makecell}
\usepackage{multirow}
\usepackage{times}
\usepackage{textcomp}
\usepackage{floatrow}
\usepackage{algorithm}
\usepackage{algpseudocode}
\usepackage[font=small,labelfont=bf]{caption}
\DeclareMathAlphabet{\mathcal}{OMS}{cmsy}{m}{n}

\newcommand*{\Comb}[2]{{}^{#1}C_{#2}}

\long\def\digest{{\em DIGEST}}

\long\def\digest{{\em DIGEST }}
\long\def\SRP{{\em SRP }}

\title{\LARGE \bf
Semantic Robot Programming for Goal-Directed Manipulation in Cluttered Scenes
}

\author{Zhen Zeng \hspace{0.5cm} Zheming Zhou \hspace{0.5cm} Zhiqiang Sui \hspace{0.5cm} Odest Chadwicke Jenkins
\thanks{Z. Zeng, Z. Zhou, Z. Sui and O.C. Jenkins are with the Department of Electrical Engineering and Computer Science, University of Michigan, Ann Arbor, MI, USA, 48109-2121
{\tt\small [zengzhen|zhezhou|zsui|ocj]@umich.edu}}
}

\begin{document}

\maketitle
\begin{abstract}
We present the Semantic Robot Programming ({\em SRP}) paradigm as a convergence of robot programming by demonstration and semantic mapping. In {\em SRP}, a user can directly program a robot manipulator by demonstrating a snapshot of their intended goal scene in workspace. The robot then parses this goal as a scene graph comprised of object poses and inter-object relations, assuming known object geometries. Task and motion planning is then used to realize the user's goal from an arbitrary initial scene configuration. Even when faced with different initial scene configurations, {\em SRP} enables the robot to seamlessly adapt to reach the user's demonstrated goal. 
For scene perception, we propose the Discriminatively-Informed Generative Estimation of Scenes and Transforms ({\em DIGEST}) method to infer the initial and goal states of the world from RGBD images. The efficacy of {\em SRP} with {\em DIGEST} perception is demonstrated for the task of tray-setting with a Michigan Progress Fetch robot. Scene perception and task execution are evaluated with a public household occlusion dataset and our cluttered scene dataset.

\end{abstract}

\thispagestyle{empty}
\pagestyle{empty}

\section{Introduction}

Many service robot scenarios, such as setting up a dinner table or organizing a shelf, require a computational representation of a user's desired world state. For example, how is the dinner table to be set, or how is the shelf to be organized.  More specifically, what are the objects involved in the task, what are the desired poses of those objects, and what are the important spatial relationships between objects. Towards natural and intuitive modes of human-robot communication, we present the {\bf Semantic Robot Programming ({\em SRP})} paradigm for declarative robot programming over user demonstrated scenes.  In {\em SRP}, we assume a robot is capable of goal-directed manipulation~\cite{Suietal2017ijrr} for realizing an arbitrary scene state in the world.  A user can program such goal-directed robots by demonstrating their desired goal scene.  \SRP assumes such scenes can be perceived from partial RGBD observations, which has proven a challenging problem in itself.

Goal-directed manipulation requires a true closing of the loop between perception and action, beyond the existing intellectual silos.  Advances in object detection~\cite{girshick2014rich, ren2015faster} from appearance has improved greatly in filtering of background noise and focused attention to objects of interest.  However, the applicability of such vision-based methods robot perception remains unclear, especially for the purposes of goal-directed manipulation.   This circumstance has given rise to new approaches to semantic mapping~\cite{kuipers2000spatial, rusu2008towards, herbst2014toward} to computationally model a robot's environment into perceivable objects with robot-actionable affordances.  

We posit semantic mapping offers a springboard to new forms of robot programming, such as Semantic Robot Programming, where semantic maps provide a generic abstraction layer for robot programming.  In our approach to this problem, we must bridge the gap of interoperation between semantic mapping and existing methods for goal-directed task planning~\cite{fikes1972strips, laird1987soar}, grasp planning~\cite{ten2016localizing} and motion planning~\cite{ompl}.  We have previously proposed methods for scene estimation~\cite{sui2015axiomatic,Suietal2017ijrr} from robot RGBD sensing that used scene graphs expressed axiomatically as a semantic mapping abstraction.  This abstraction allowed for ready use with modern task, grasp, and motion planning systems.  The resulting of closing this loop with a semantic abstraction layer is envisioned to enable portable robot-executable expressions accessible across a variety of modalities, including: natural language, visual programming, and put-that-there gesturing~\cite{cannon1993point,kemp2008point}.  However, the computational cost of inference over scenes is asymptotically intractable as the number of objects grows.

\begin{figure}[t]
\centering
  \includegraphics[width=\linewidth]{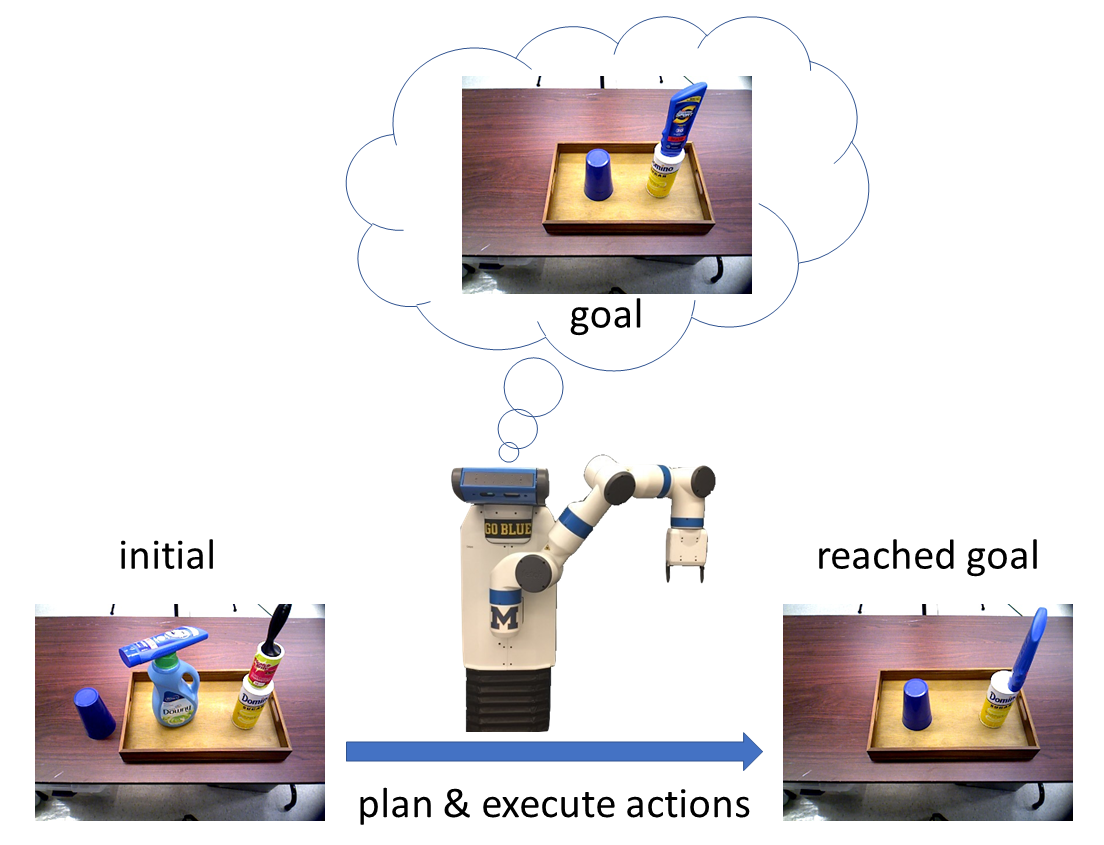}
   \caption{\footnotesize A robot preparing a tray through goal-directed manipulations. Given the observation of the user desired goal state and the initial state of the tabletop workspace, the robot first perceives the axiomatic scene graph of the goal and initial state, and then plan and execute goal-directed actions to prepare the tray the way the user desires.}
\label{fig:illustration}
\end{figure}

In this paper, we propose the paradigm of Semantic Robot Programming for robot manipulators with a complementary method for more tractable scene perception.  \SRP is a declarative approach to programming robots through demonstration, where users only need to demonstrate their desired state of the world.  \SRP is general across methods of perception, given the perceived scene is represented axiomatically.  For scene perception, we present the Discriminatively-Informed Generative Estimation of Scenes and Transforms ({\em DIGEST}) method to infer the initial and goal scene states for \SRP from RGBD images.  {\em DIGEST} brings together discriminative object detection and generative pose estimation for inference of 6 DOF object poses in cluttered scenes, assuming the number of objects is known.
Given perceived initial and goal scenes, the robot can plan and execute goal-directed manipulation to autonomously transit the world from the initial to the goal state.

We evaluate the \SRP paradigm in tray-setting task scenario with the Michigan Progress Fetch robot (Figure~\ref{fig:illustration}). We benchmark the performance of {\em DIGEST} on a household occlusion dataset~\cite{aldoma2012point} and our cluttered scene dataset. We demonstrate that \SRP is effective in understanding the goal of a task given a demonstrated snapshot of the goal scene. And, the robot is able to plan and execute goal-directed manipulation actions to reach the goal from various initial states of the world. We additionally found {\em DIGEST} performs favorably in comparison with state-of-the-art methods for scene perception, such as D2P~\cite{Narayanan-RSS-16}, with fewer assumptions of prior knowledge.

\section{Related Work}

\SRP builds on much existing work in robot Programming by Demonstration (PbD) and scene perception for manipulation. Similar to robot PbD, \SRP aims to enable users to effectively communicate their objectives to robots for performing manipulation tasks. We posit advances in scene perception for manipulation offers new avenues for extending the ease and intuitiveness of robot PbD.

\subsection{Programming by Demonstration}
To improve communication of tasks from a user to a service robot, existing research has focused on learning low-level skills from users. Different approaches have been proposed in Programming by Demonstration (PbD) for low-level learning of skills, such as trajectories~\cite{nakanishi2004learning}~\cite{akgun2012keyframe} and control policy~\cite{chernova2009interactive}~\cite{grollman2010incremental} in robot \textit{configuration space}. These methods are inherently limited to world states in \textit{workspace} that are similar to the ones in the demonstrations. By representing the goal of a task in the \textit{workspace} instead of in the \textit{configuration space}, goal-directed manipulation can reason and plan its actions to reach the goal from arbitrary initial world states.

Other work has focused on the high-level aspects of a task. Veeraraghavan et al.~\cite{veeraraghavan2008teaching} propose learning high level action plan for a repetitive ball collection task from demonstrations. Ekvall et al.~\cite{ekvall2008robot} focus on learning task goals and use a task planner to reach the goal. Chao et al.~\cite{chao2011towards} provide an interface for the user to teach task goals in a tabletop workspace. However, these methods wind up simplifying the scene perception problem by using planar objects, box-like objects or objects with distinguishing colors, that are far from real world scenarios. Recently, Yang et al.~\cite{yang2015robot} have proposed learning action plans in real world scenario, similar to our robot programming paradigm that works with real world objects.

\subsection{Scene Perception for Manipulation}
Being able to perceive objects in real world scenarios and act on them remains a challenge. Some works are able to extract grasping point~\cite{ciocarlie2014towards, lenz2015deep, ten2015using} in point cloud data, however, their methods do not provide a structural understanding of the scene, failing to support goal-directed manipulation on objects.

Although not directly targeted at scene perception for manipulation, work on object pose estimation are highly related to our work. Feature-based object pose estimation methods suh as spin images~\cite{johnson1999using}, FPFH~\cite{rusu2009fast}, OUR-CVFH~\cite{aldoma2012our} and VFH~\cite{rusu2010fast}, rely on feature matching between the object model and observation, however, the problem is that the performance of feature-based methods degrades as the environment becomes more cluttered and key features are occluded. Recently, Narayanan et al. proposed D2P~\cite{Narayanan-RSS-16}, which outperforms feature-based method OUR-CVFH on the household occlusion dataset~\cite{aldoma2012point}. D2P renders multiple scene hypotheses, and use A* to search for the hypothesis that best explains the observation. In our experiments, we demonstrate that our proposed scene estimation method {\em DIGEST} outperforms D2P on the household occlusion dataset.

To plan goal-directed manipulations, knowing the object poses is not sufficient, however. The robot must have a structural understanding of the scene, that is, the inter-object spatial relations. Given observations of the scene, our work estimates a scene graph that represent the scene structure. Liu et al.~\cite{liu2015table} also estimate a scene graph given observations, however, their approach approximates objects as oriented bounding boxes. Sui et al. proposed a generative approach (AxMC)~\cite{Suietal2017ijrr} for scene graph estimation and use Markov Chain Monte Carlo (MCMC) to search for the best scene graph hypothesis that explains the observations.

Both D2P and AxMC assume that the robot knows what objects are present in the scene, and objects are standing in their upright poses, thus both methods can only estimate 3 DOF poses of objects (i.e., $x, y, \theta$). However, these assumptions are too strong in real world scenarios. Instead, our scene estimation method {\em DIGEST} does not rely on any of these assumptions, and it can estimate 6 DOF poses of objects, as long as the number of objects in the scene is known.

\begin{figure*}[t]
  \centering
  \includegraphics[width=\linewidth]{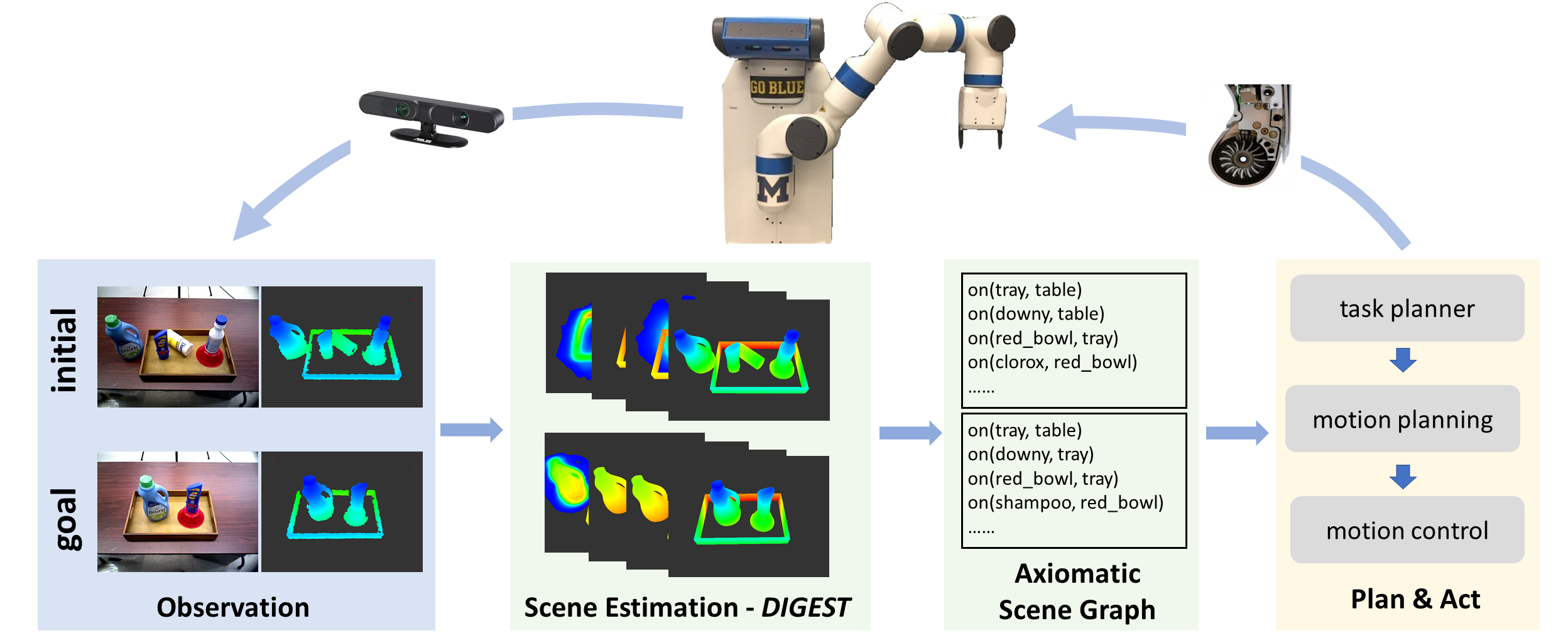}
  \caption{\footnotesize Our goal-directed robot programming has three stages: 1) Given the RGB-D observation of the goal and initial scene, we use the proposed scene estimation method {\em DIGEST} to detect object and estimate the 6 DOF pose of objects; 2) Axiomatic scene graphs can be derived from the estimated object poses, which express the inter-object spatial relations; 3) By describing the goal and initial scene graph by PDDL, the robot uses a task planner (e.g., STRIPS) to plan and execute a sequence of goal-directed actions to reorganize the objects in the scene, reaching the same inter-object relations in the goal scene graph.}
  \label{fig:pipeline}
\end{figure*}

\section{Problem Statement}
\SRP with \digest assumes the number of objects $N_c$ present in the scene, 3D mesh models $\mathbf{M}=\{m_1,\cdots,m_l\}$ for a set of objects. The robot is assumed capable of performing a set of manipulation actions $\mathbf{A}=\{a_1,\cdots,a_n\}$ with known pre-conditions and post-conditions on these objects. We assume as given RGB-D observation of the goal scene $o_G$ specified by the user at time $t$, and the current scene $o_I$ at a later time $t+T$. The objective of \SRP is to plan a sequence of goal-directed manipulation actions $\{a_i,\cdots,a_j\}$ to rearrange objects in the world such that the inter-object relations in $s_G$ are satisfied; where \digest infers the goal scene graph $s_G$ and the initial scene graph $s_I$, respectively.

We use a list of axiomatic assertions to describe a scene as a scene graph. The scene state at time $t$ is expressed as a scene graph $s_t=\{v^i(x_t)\}_{i=0}^K$, where $v^i \in \{exist, clear, on, in\}$ is an axiomatic assertion parameterized by $x_t=\{q^j_t\}_{j=0}^{N_c}$, with $q^j_t$ denoting the pose of $j$th object at time $t$, $N_c$ being the number of objects, and $K$ being the total number of axiomatic assertions. In our work, the assertions are limited to spatial relations that can be tested geometrically. The 6 DOF pose $q^j_t=[x^j_t,y^j_t,z^j_t, \phi^j_t, \psi^j_t, \theta^j_t]$ of each object is estimated, consisting 3D position ($x^j_t, y^j_t, z^j_t$) and orientation ($\phi^j_t, \psi^j_t, \theta^j_t$). The scene graph can be inferred from the estimated object poses, as explained later in Section \ref{sec:scene_graph_structure}.

\section{Methods}
The \SRP paradigm consists of the perception of goal and initial scene states, and  the planning and execution stages, as shown in Figure \ref{fig:pipeline}. 
Given observations of a cluttered scene, the generative sampling inference process over object poses is informed by detections from a discriminative object detector. A scene graph encoding inter-object relations is geometrically inferred from an estimate of inferred object poses.  The resulting scene graph is then expressed axiomatically for use in task planning and execution.

\subsection{DIGEST Cluttered Scene Estimation}
\begin{figure*}[t]

\includegraphics[width=0.9\linewidth]{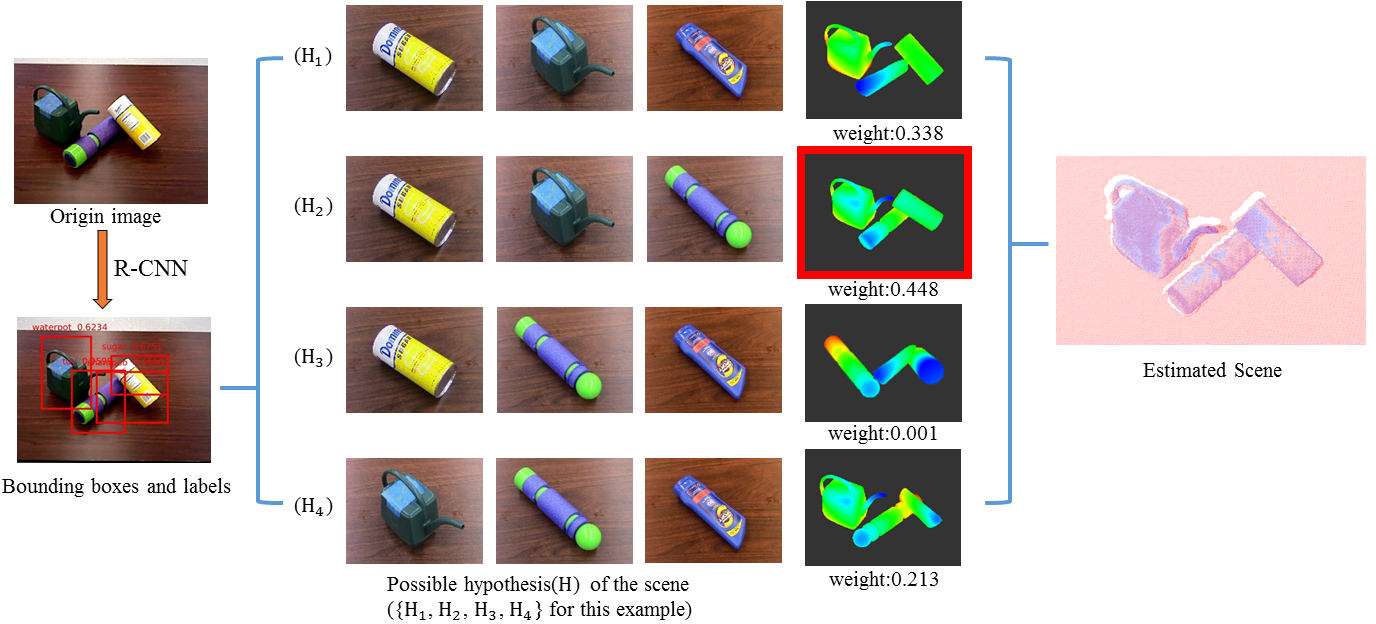}

\caption{\footnotesize The proposed \digest method for cluttered scene estimation. First, the observed RGB image is passed through a R-CNN object detector trained on our grocery object dataset. The R-CNN object detector outputs a set of bounding boxes, with associated object label and detection confidence. Knowing the number of object present in the scene, possible scene hypotheses are enumerated, e.g.,  $\Comb{4}{3}=4$ scene hypotheses are generated in this example. For each scene hypothesis, particle filtering is applied to estimate object poses that best explains the observed depth. After convergence, {\em DIGEST} outputs the estimated object poses for the most likely scene hypothesis.}
\label{fig:framework}
\end{figure*}

Given observed RGB-D image pair of a cluttered scene at time $t$, the objective is to estimate the object poses $q_t^j, j=1,\cdots,N_c$. We utilize the discriminative power of a pre-trained object detector to first obtain a set of bounding boxes with object labels. These bounding boxes are used to guide the generative process of scene hypotheses sampling. An overview of the cluttered scene estimation is as illustrated in Figure \ref{fig:framework}.

\subsubsection{Object Detection and Scene Hypotheses Generation}\label{sec:object_detection_method}
Given an RGB image, $m$ bounding boxes are detected by the object detector. We use $B_i$ ($0 \leq i \leq m$) to denote the bounding box. In the output of the object detector, each $B_i$ is associated with a list of object detection confidence $v(L_j | B_i)$, where $L_j$ is the object class. For each $B_i$, we generate an object candidate $C_i$,
\begin{equation} \label{eq:candidate}
C_i = \{ \underset{L_j}{\arg\max} \, v( L_j | B_i ), \, B_i \}
\end{equation}
which is a set including the object label with the highest confidence measure and the associated bounding box.
For $m$ generated candidates, the number of scene hypotheses $h$ equals to $N_c$ chooses $m$, i.e.,
\begin{equation}
    h =
    \begin{cases}
      \Comb{m}{N_c}, & \text{if}\ N_c \leq m \\
      1, & \text{otherwise}
    \end{cases}
\end{equation}
Thus, if the number of candidates is greater or equal to the number of objects in the scene, each scene hypothesis ${H_i}$ contains a combination of $N_c$ candidates selected from $m$ candidates. If the number of candidates is less than $N_c$, just one scene hypothesis with $m$ candidates will be generated. 

\subsubsection{Bootstrap Filtering for Pose Estimation}\label{sec:bootstrap_filter_method}
Each scene hypothesis ${H_i}$ is modeled as a random state variable $x_t$, composed of a set of real-valued object poses. Object poses are assumed to be statistically independent.  We model the inference of the state from robot observation as a Bayesian filter problem. Compared to traditional Bayesian filter problems, we have only one observation: a snapshot of the scene instead of a history of observations. Thus, we apply Iterated Likelihood Weighting~\cite{mckenna2007tracking} to bootstrap the scene estimation process, where $z_1=z_2=\cdots=z_t$ and the state transition in the action model is replaced by a zero-mean Gaussian noise. We approximate the belief distribution by a collection of $N$ particles $\{x^{(j)}_t$ weighted by $w^{(j)}_t\}^{N}_{j=1}$,
\begin{equation}\label{eq:particle}
p( x_{t} | z_{1:{t}} )
   \propto 
p( z_{t} | x_{t} )\sum_{j} w^{(j)}_{t-1}p( x_{t} | x^{(j)}_{t-1}, u_{t-1}) 
\end{equation}
\begin{equation}
x_t^{(j)} \sim \sum_{j} w^{(i)}_{t-1}p( x_{t} | x^{(i)}_{t-1}, u_{t-1})
\end{equation}
as described by \cite{Dellaert_MCL}. To evaluate the weight $w_t^{(j)}$ for particle $x_t^{(j)}$, we render a depth image based on the object poses in $x_{t}^{(j)}$, and compare it against the observed depth image $\hat{z}_{t}^{(j)}$,
\begin{equation}
    w_t^{(j)} = e^{- \lambda_r \cdot \mathrm{d}(z, \hat{r}_{t}^{(j)})}
\end{equation}
\noindent where $\lambda_r$ is a constant scaling factor. $\mathrm{d}(z,\hat{r}_{t}^{(j)})$ is the sum of the Euclidean distance between the 3D points projected back from depth images $z$, $\hat{r}_{t}^{(j)}$, using the intrinsic parameters of the camera. Pose estimation is performed over successive iterations that:
1) compute the weight of each particle, 2) normalize the weights to one, 3) draw $N$ particles by importance sampling, and 4) diffuse each sampled particle by a zero-mean Gaussian noise. After maximum number of iterations, the most likely particle as the scene estimate for scene hypothesis $H_i$:

\begin{equation}
x_t = \underset{{x}_t^{(j)}} {\arg\max} \, p(x_t^{(j)}|z_{1:t})
\end{equation}

\subsubsection{Final Scene Ranking}
After particle filtering for all scene hypotheses, we have a scene estimate $x_t$ for each scene hypothesis. We then rank them based on the likelihood of each $x_t$ as computed earlier. The most likely $x_t$ is taken as the scene estimate and is then used to derive the scene graph.

\begin{figure*}[t]
\centering
  \includegraphics[width=0.9\linewidth]{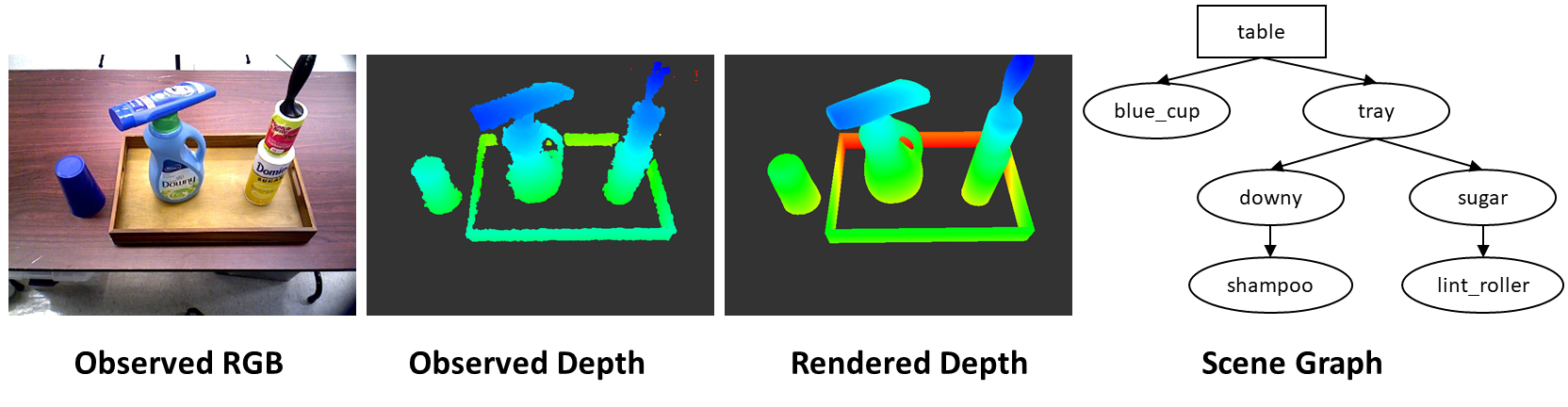}
	\caption{\footnotesize An axiomatic scene graph example. In the scene graph derived from the estimated object poses, each node corresponds to an object, and each edge indicates the supporting relation between objects. 	\textit{table} is by default the root node.}
\label{fig:scene_graph_pddl}
\end{figure*}

\subsection{Scene Graph Structure}\label{sec:scene_graph_structure}
The objects pose estimation of a cluttered scene can be turned into an axiomatic scene graph. We use following axiomatic assertions: $exist(q^j)$ for the assertion that object $j$ exists in the scene with pose $q^j$; $clear(q^i)$ for the assertion that the top of object $i$ is clear and no other objects are stacked on it; $on(q^i,q^j)$ for the assertion that object $i$ is stacked on object $j$; $in(q^i,q^j)$ for the assertion that object $i$ is in object $j$. An example of a scene graph is given in Figure \ref{fig:scene_graph_pddl}. 

To assert the proximity relations between two objects $i, j$, we add a \textit{virtual object} $q^\gamma$ with geometry $m^\gamma$ into the scene graph, with $m^\gamma$ being a shape that can be arbitrarily defined based on the application, and $q^\gamma$ being the identity pose in the frame of object $i$. Then, the proximity relation between objects $i,j$ can be encoded by $\{has(q^i,q^\gamma),\ in(q^\gamma,q^j)\}$, where $has(q^i,q^\gamma)$ asserts that object $i$ has a \textit{virtual object} $q^\gamma$ attached to its frame.
When the parent object $i$ is in a new location, the robot can adapt to the new scenario by placing the child object $j$ within the region of $m^\gamma$ attached to the frame of $i$.

To determine the stacking relations between the objects, we use simple heuristics. In the 3D mesh object models, the z-axis of each object is the gravitational axis when the object stands upright. The dimensions $\{h_x, h_y, h_z\}$ of the 3D box that encloses each object model are given as prior knowledge. In order to determine whether object $i$ is being supported by another object, two heuristics are tested: (1) if one of the object axes (e.g., x-axis) is aligned with the gravitational axis, then the height $h_i$ of the 3D volume occupied by the object equals to the corresponding dimension (e.g. $h_x$) of the provided 3D enclosing box. A simple rule $z^i - h^{table} > 0.5h^i$ is used to determine whether object $i$ is being supported by another object; (2) if none of the object axes are aligned with the gravitational axis, then object $i$ is being supported by another object.

The set of objects that is being supported by other objects is sorted with increasing $z$ values of the object pose, and is denoted as $O_{\text{s}}$, the remaining objects are denoted as $O_{\text{r}}$. For each object $i \in O_{\text{s}}$, a heuristic measure is used to determine which object $j \in O_{\text{r}}$ is supporting $i$,
\[ \operatorname{arg\,max}_j f(r_b(q^i),r_t(q^j)) \]
where $f(r_1,r_2)$ measures the overlapping area of two regions $r_1, r_2$, and $r_t(q^i), r_b(q^i)$ represent the projected region on the table of the top and bottom surface of object $i$, respectively. Once the supporting object for $i \in O_{\text{s}}$ is identified, $i$ is moved from set $O_{\text{s}}$ to $O_{\text{r}}$. With the supporting relation between a pair of objects $i, j$ identified, the corresponding axiomatic assertion is expressed as either $on(q^i, q^j)$ or $in(q^i, q^j)$, depending on the geometry type of the supporting object $j$ being convex or concave.

\section{Implementation}

\subsection{RCNN object detector}
We employ R-CNN~\cite{girshick14CVPR} as our discriminative object detector as described in section \ref{sec:object_detection_method}. R-CNN first generates object bounding boxes given an image, then for each bounding box, it outputs the confidence measure through a deep convolutional neural network. For the sake of efficiency and performance, we replace the original selective search~\cite{uijlings2013selective} with EdgeBox~\cite{zitnick2014edge} for object proposal generation. We train an R-CNN object detector on our object dataset that includes 15 grocery objects. The dataset contains 8366 ground truth images (\texttildelow 557 average ground truth images for one object) and 60563 background images. We fine tuned our object detector on a pre-trained model on ImageNet~\cite{imagenet_cvpr09}. 

\subsection{Particle Filtering and parallelization}
During bootstrap filtering for pose estimation as described in section \ref{sec:bootstrap_filter_method}. Each object in each particle $x_t^{(j)}$ is initialized by candidate $C_i$ in the scene hypothesis, the object label $l_i$ determines which 3D mesh object model to use, and the initial pose is uniformly sampled inside the bounding box $B_i$. A parallel graphics engine rapidly renders depth images given all particles. CUDA is used to compute the weights of all particles in parallel. Through our experiment, we fix particle filter iteration to 400 and use 625 particles.

In the particle filtering process, the pose of each object is estimated sequentially. For example, if there are four hypothesized objects and 400 particle filter iterations, the pose of the object with the maximum detection confidence is estimated in the first 100 iterations. Then the pose of the object with the 2nd largest detection confidence is estimated in the next 100 iterations, with the first object fixed at the most likely pose. We carry on the estimation process iteratively for the remaining objects.

\subsection{Planning and Execution}
Given the observation of the goal state of the world, the robot estimates the goal scene graph, and stores the desired inter-object relations by PDDL~\cite{mcdermott1998pddl}. Similarly, the robot estimates and stores the initial inter-object relations by PDDL. With sets of PDDL that describe the initial and goal state, the robot uses a task planner to plan a series of goal-directed actions to rearrange objects in the initial scene, such that the same inter-object relations in the goal scene graph are satisfied. We use breadth first search STRIPS\cite{fikes1971strips} as our task planner. Note that the robot does not need to rearrange the objects with the exact same poses as in the goal scene, as long as the same inter-object relations are achieved, similarly to how human would arrange a set of daily objects based on simple instructions.

The task planner gives a sequence of high-level pick-and-place actions. To pick an object, the robot is given a set of pre-computed grasp poses of the object using~\cite{ten2015using}, and uses Moveit!~\cite{sucan2013moveit} to check which grasp pose it can generate a collision-free trajectory for, and use that for grasping. To place an object, the robot sample place poses in the empty space that satisfies the desired inter-object relations, and again use the place pose it can generate a collision-free trajectory for.

\section{Experiments}

In our experiments, we first evaluate our scene estimation method on a public household occlusion dataset and our cluttered scene dataset, and then evaluate our overall semantic robot programming paradigm in tray setting tasks. {\em DIGEST} outperforms the state of the art method D2P on the household occlusion dataset, and outperforms FPFH on our cluttered scene dataset. We demonstrate the effectiveness of our system for programming a robot to complete various tray-setting tasks through goal-directed manipulations. We run all experiments on a computer with an Titan X Graphics card and CUDA 7.5.

\subsection{DIGEST: Cluttered Scene Estimation}
To evaluate {\em DIGEST} on pose estimation, we benchmarked the performance of {\em DIGEST} on two different datasets: household occlusion dataset~\cite{aldoma2012point}, and our cluttered scene dataset. The household occlusion dataset contains objects standing up right, thus it only affords benchmarking on 3 DOF object pose estimation. In our cluttered scene dataset, objects can be in arbitrary pose, and we use it for benchmarking on 6 DOF object pose estimation. Object pose estimation accuracy is calculated as the percentage of correctly localized objects over the total number of objects in the dataset. An object is correctly localized if the pose error falls within certain position error threshold $\Delta t$ and rotation error threshold $\Delta \theta$. The position error is the Euclidean distance error in translation; the rotation error is the absolute angle error in orientation. For rotationally symmetric objects, the rotation error about the symmetric axis is ignored.

\begin{figure}[!t]
\includegraphics[width=\columnwidth]{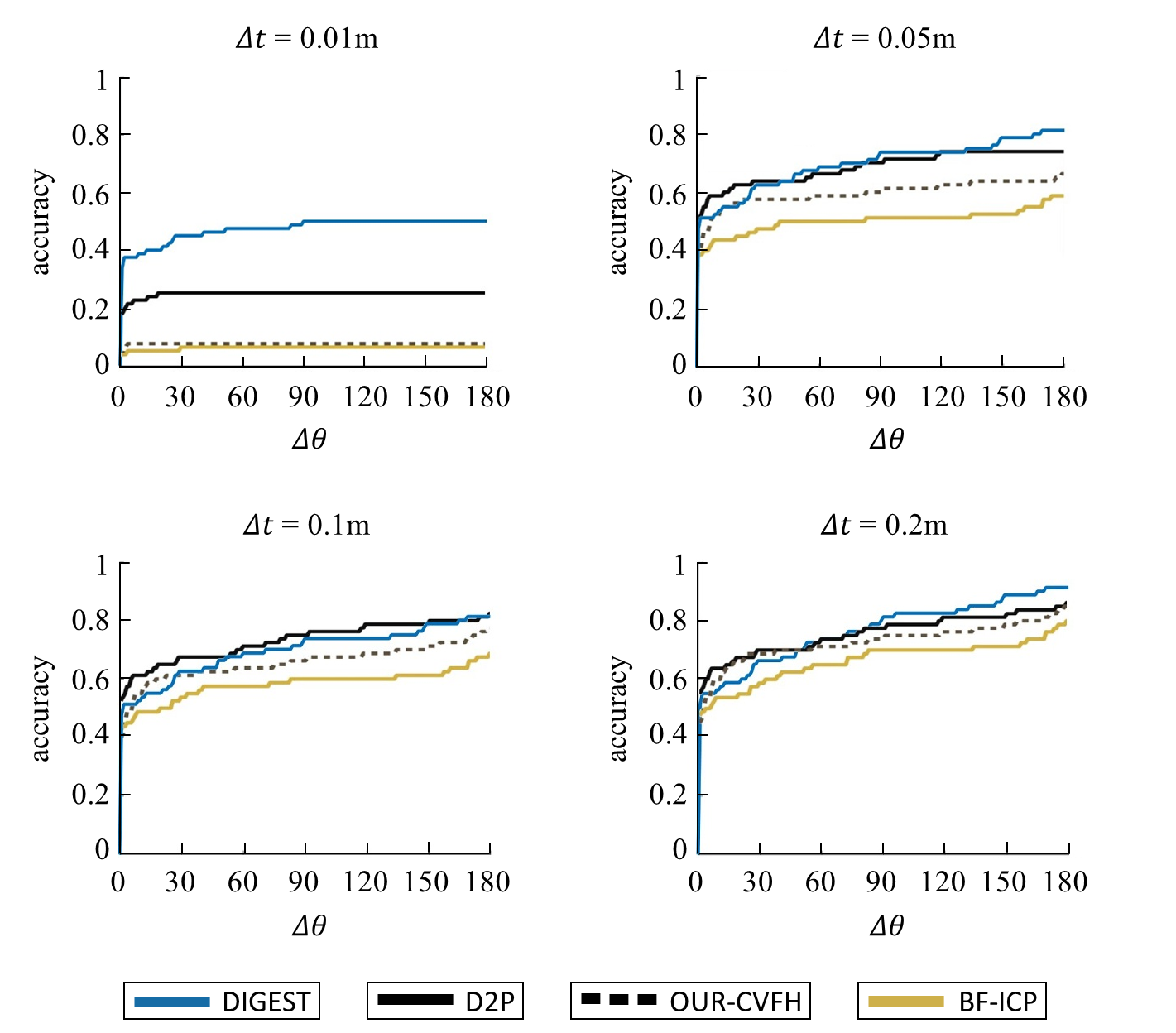}

\caption{\footnotesize Object pose estimation benchmark of \digest on public household object dataset~\cite{aldoma2012point}, compared with three baseline methods: D2P, OUR-CVFH and BF-ICP for different correctness criteria $\Delta t$, $\Delta \theta$. {\em DIGEST} outperforms D2P for strict correctness criteria, and performs on par with D2P for relaxed correctness criteria.}
\label{fig:d2p_progress_exp_pose_err}
\end{figure}

\subsubsection{Household Occlusion Dataset -- 3 DOF Object Poses}
The household occlusion dataset contains 22 test scenes with 80 objects in total. The test scenes include objects such as milk bottles, laundry items, mugs and etc; We compare \digest against three baseline methods as described in ~\cite{Narayanan-RSS-16}, that is, D2P, OUR-CVFH~\cite{aldoma2012our}, and Brute Force ICP (BF-ICP). D2P also uses an R-CNN object detector as part of their pose estimation process, but it is not clear what hyper parameters they choose during the training phase of the object detector. In order to avoid bias in the training of the object detector, we use their object detector on the household occlusion dataset.

\begin{figure}[!t]
\includegraphics[width=\columnwidth,]{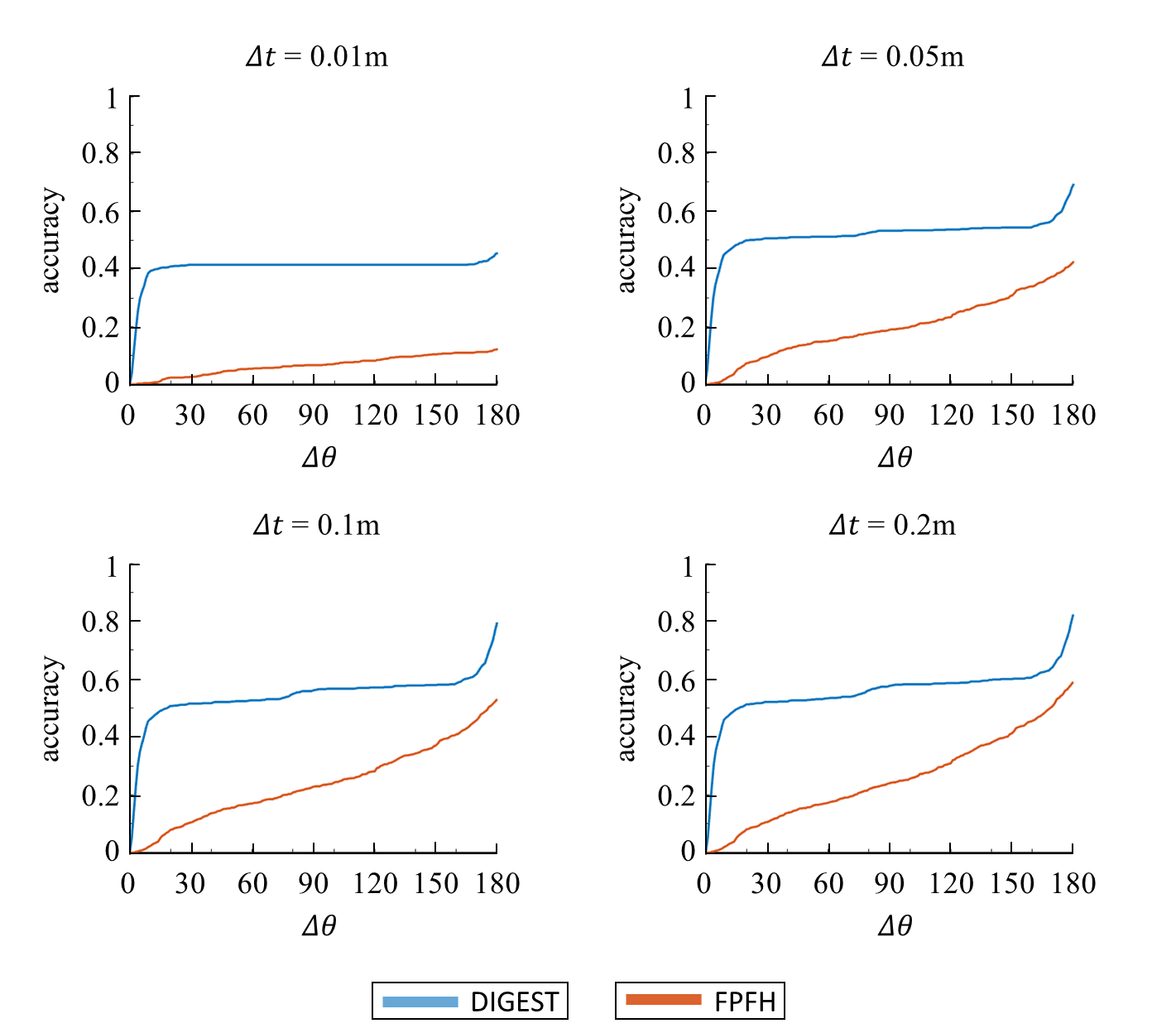}

\caption{\footnotesize Object pose estimation benchmark of \digest on our cluttered scene dataset, compared with baseline method FPFH under different correctness criteria $\Delta t$, $\Delta \theta$. {\em DIGEST} outperforms FPFH with large margin.}
\label{fig:digest_fpfh}
\end{figure}

When only little error is allowed for an estimated pose to be counted as correct, as shown in the left upper plot in Figure \ref{fig:d2p_progress_exp_pose_err}, the accuracy of \digest is nearly twice the accuracy of D2P. As we relax the tolerance on the pose estimation error, as shown in the other three plots in Figure \ref{fig:d2p_progress_exp_pose_err}, \digest performs on par with D2P. Overall, \digest outperforms D2P since (1) \digest explores the state space a lot more than D2P, as we do not discretize the state space, and (2) \digest does not use ICP for local search, which D2P employs for pose estimation. In terms of run time, \digest takes around 30 seconds (varying with the number of objects and the size of object mesh), which is faster than 139.74 seconds reported in D2P.

\subsubsection{Cluttered Scene Dataset -- 6 DOF Object Poses}
We collect a cluttered scene dataset with 16 different sceness, and 72 objects in total. This dataset includes laundry items, kitchen items and toy with non-trivial geometry. The number of objects in each scene ranges from 3 to 7. This dataset is much more challenging than the household object dataset, as the objects can have random 6 DOF poses. We compare the performance of {\em DIGEST} with FPFH~\cite{rusu2009fast}, as shown in Figure \ref{fig:digest_fpfh}.

\begin{figure*}
\includegraphics[width=1.0\linewidth]{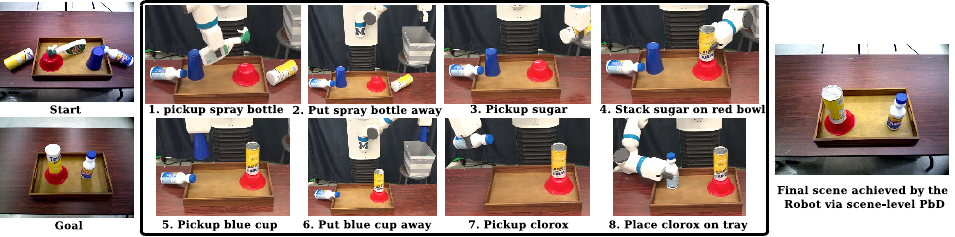}
\caption{\footnotesize Our robot performing goal-directed manipulation (middle columns) to prepare a tray (right) satisfying the user-demonstrated goal (left bottom).}
\label{fig:manipulation_exp}
\end{figure*}
\subsection{Semantic Robot Programming: Tray Setting}

We designed our experiments around a service robot scenario, as illustrated in Figure \ref{fig:illustration}. The robot needs to prepare a tray as specified by the user int the goal scene. We tested our system on scenes of 4 to 6 objects including the tray, with different inter-object relations, such as stacking and proximity relations. The robot is able to perceive the initial and goal state, then plan and execute goal-directed actions to satisfy the inter-object relations in the goal scene graph. 

\begin{figure*}[t!]
    \includegraphics[width=\linewidth]{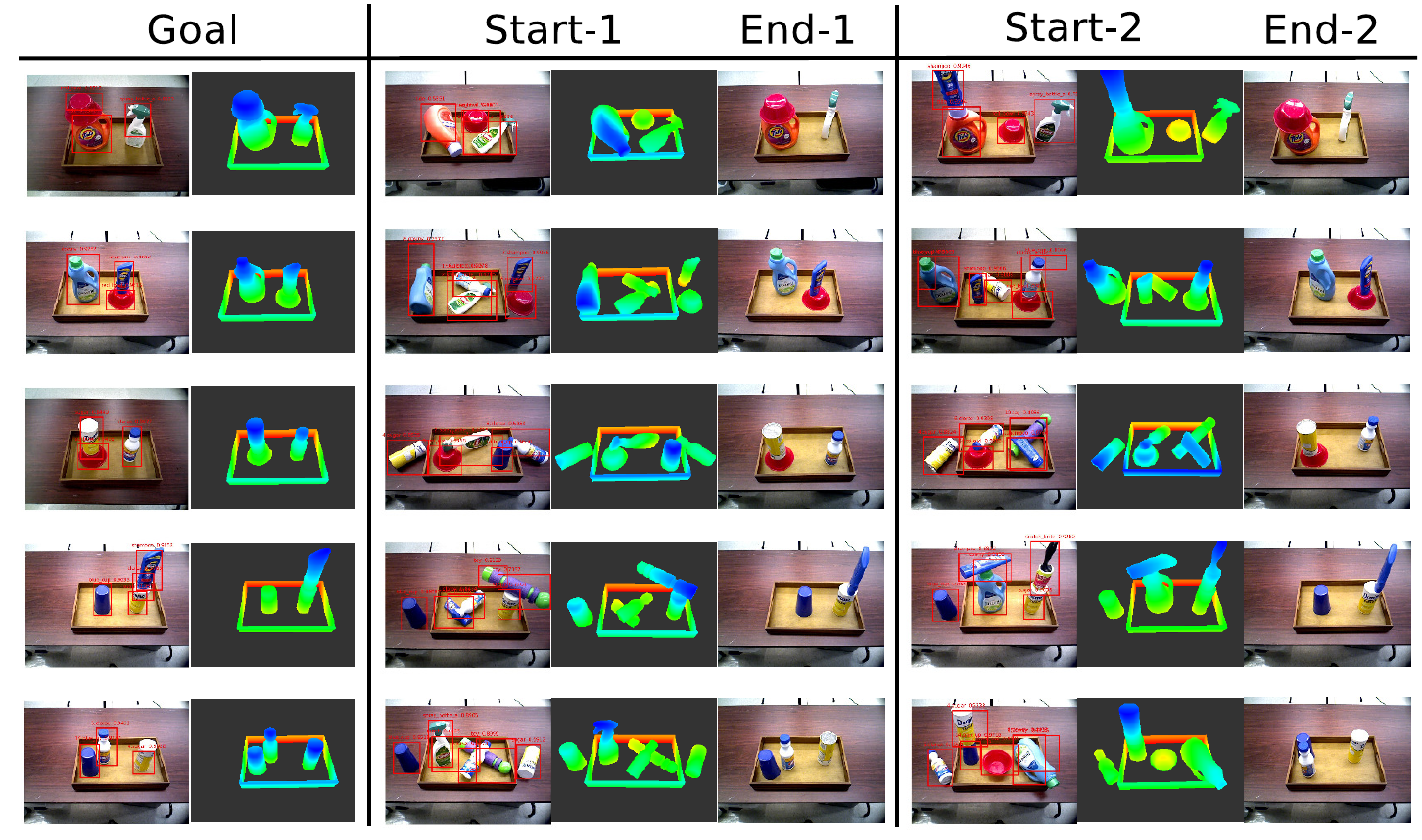}
  \caption{\footnotesize Our \SRP system tested for 5 different tray preparation tasks. The left column shows the goal scene. For each goal, the robot starts from two different initial states and successfully performs goal-directed manipulations to prepare the tray. Depth images are rendered based on 6 DOF object poses output by {\em DIGEST}.}
  \label{fig:table_digest}
\end{figure*}

An example of \SRP for goal-directed manipulation is shown in Figure \ref{fig:manipulation_exp}.  Based on the scene graph inferred from the object pose estimates, the robot generates a sequence of goal-directed actions to achieve the goal state.  Our tray setting experiments are shown in Figure \ref{fig:table_digest}, and more detail in the accompanying video\footnote{\url{http://progress.eecs.umich.edu/projects/srp}}. The goal and start scenes are well estimated as a collection of 6DOF poses of objects.  The robot successfully sets up a tray as the user desired in 10 out of 10 different tray setting experiments.

\section{Conclusion}
We have presented Semantic Robot Programming as a paradigm for users to easily program robots in a declarative goal-directed manner. We demonstrate the effectiveness of \SRP using the proposed \digest scene perception method on two datasets of objects in occlusion and clutter: both house occlusion dataset and our cluttered scene dataset. 
Through our approach to generative-discriminative perception, \SRP with \digest is able to perceive, reason, and act to realize an arbitrary user-demonstrated goal in cluttered scenes.

There are many future directions to pursue, such as motion planning over sequences of general manipulation actions.  Currently, grasp point localization~\cite{ten2015using} is used to select good grasp poses for object picking. However, such selected grasp poses are not necessarily appropriate for a later placement actions.  Visual inspection on selected grasps is done before robot execution. Ideally, appropriate grasp poses would be provided by a manipulation affordance mechanism, such as Affordance Templates~\cite{hart2015affordance} associating robot action with an object. Such affordance mechanisms would allow for investigation of more flexible task and motion planning over sequences of actions.  We further posit scene perception can be made to run in interactive-time through a thoughtful parallelized implementation, enabling potentially interactive planning and manipulation execution.

\bibliographystyle{abbrv}

\end{document}